\documentclass{article}

\usepackage{arxiv}

\usepackage[utf8]{inputenc} 
\usepackage[T1]{fontenc}    
\usepackage{hyperref}       
\usepackage{url}            
\usepackage{booktabs}       
\usepackage{amsfonts}       
\usepackage{nicefrac}       
\usepackage{microtype}      
\usepackage{graphicx}
\usepackage{doi}
\usepackage{xcolor}
\usepackage{subcaption}

\title{Exploring the Feasibility of Generating Realistic 3D Models of Endangered Species Using  DreamGaussian: An Analysis of Elevation Angle's Impact on Model Generation}

\author{{\hspace{1mm}Selcuk Anil Karatopak} \\
	Huawei Türkiye R\&D Center \\
    Istanbul, Türkiye \\
	\texttt{selcuk.anil.karatopak@huawei.com} \\
	\And
	{\hspace{1mm}Deniz Sen} \\
	Huawei Türkiye R\&D Center \\
    Istanbul, Türkiye \\
	\texttt{deniz.sen1@huawei.com} \\
}

\renewcommand{\shorttitle}{\textit{arXiv} Template}

\hypersetup{
pdftitle={A template for the arxiv style},
pdfsubject={q-bio.NC, q-bio.QM},
pdfauthor={David S.~Hippocampus, Elias D.~Striatum},
pdfkeywords={First keyword, Second keyword, More},
}

\begin{document}
\maketitle

\begin{abstract}
    Many species face the threat of extinction. It's important to study these species and gather information about them as much as possible to preserve biodiversity. Due to the rarity of endangered species, there is a limited amount of data available, making it difficult to apply data requiring generative AI methods to this domain. We aim to study the feasibility of generating consistent and real-like 3D models of endangered animals using limited data. Such a phenomenon leads us to utilize zero-shot stable diffusion models that can generate a 3D model out of a single image of the target species. This paper investigates the intricate relationship between elevation angle and the output quality of 3D model generation, focusing on the innovative approach presented in DreamGaussian. DreamGaussian, a novel framework utilizing Generative Gaussian Splatting along with novel mesh extraction and refinement algorithms, serves as the focal point of our study. We conduct a comprehensive analysis, analyzing the effect of varying elevation angles on DreamGaussian's ability to reconstruct 3D scenes accurately. Through an empirical evaluation, we demonstrate how changes in elevation angle impact the generated images' spatial coherence, structural integrity, and perceptual realism. We observed that giving a correct elevation angle with the input image significantly affects the result of the generated 3D model. We hope this study to be influential for the usability of AI to preserve endangered animals; while the penultimate aim is to obtain a model that can output biologically consistent 3D models via small samples, the qualitative interpretation of an existing state-of-the-art model such as DreamGaussian will be a step forward in our goal.
\end{abstract}


\section{Introduction}
Recently, the profound impact of Artificial Intelligence (AI) and deep learning technologies has significantly transformed our daily lives, altering a wide array of sectors in the industry. Within this ever-changing environment, generative AI has become a focal point of technological innovation due to its capability to autonomously generate realistic and intricate content. Arguably the wave of interest in AI can be partly attributed to the introduction of the stable diffusion (SD)\cite{rombach2022high} algorithm, a groundbreaking development that has pushed the state-of-the-art to seemingly impossible levels.

As a result, there has been noteworthy progress in 3D generation capabilities, expanding the potential applications of AI across different domains. With the exponential increase in data and advancements in neural implicit 3D representations, the industry has started incorporating generative AI in many different use cases. However, achieving consistent 3D representations demands a substantial amount of data samples, which may not always be available, especially in terms of clean image format.

Many species face the threat of extinction, which harnesses the importance of preserving biodiversity. As a result of the rare occurrence of these endangered species, the sparsity of related data occurs, which makes generative AI inapplicable to the domain, due to the requirement of extensive amounts of data. In this paper, we explore the possibility of combining 3D generative AI and wildlife conservation, presenting the results of current zero-shot 3D model generation approaches designed to overcome the data scarcity issue. Utilizing state-of-the-art techniques in generative novel view synthesis and implicit neural 3D representations, we aim to obtain consistent 3D models of endangered animals that suffer from the data scarcity issue. This paper seeks to not only discuss the effectiveness of 3D generative AI models under small sample situations but also to help preserve the fragile biodiversity of our planet.

\section{Related Work}

\subsection{Diffusion Models}

Much like generative adversarial networks (GANs) \cite{goodfellow2020generative}, the primary goal of the stable diffusion algorithm is to learn a transformation between the Gaussian distribution and a distribution of images. In contrast to GANs and techniques such as variational autoencoders (VAEs), which perform the transformation in a single step, SD models eliminate noise by passing the intermediate outputs from an autoencoder in many iterations. Consequently, the refined denoising process produces lifelike images, avoiding inherent issues like mode collapse. Denoising diffusion probabilistic models (DDPMs), a fundamental approach, instruct the model to reverse a forward Markov chain where the input image undergoes iterative perturbation with noise sampled from a Gaussian distribution \cite{ho2020denoising}. This results in both a forward and a backward Markov chain to transform a noise distribution into actual image data. Similar to GANs and VAEs, sampling from a Gaussian distribution and iteratively applying diffusion into the noise will result in completely new image data. It's worth noting the existence of the Latent Diffusion Model (LDM), which substantially reduces the computational cost of applying denoising over the pixel space by instead denoising in the latent space, which is considerably smaller than the image space\cite{rombach2022high}. Furthermore, in Score-based Generative Models(SGM), during the perturbation of the image in the forward chain, a score is estimated to maintain the noise level; subsequently, during the diffusion inference, the model tries to decrease this score, which as a result should give a  noiseless image output \cite{song2020score}. These methodologies form the basis for other diffusion-based generation models, applicable to both 2D novel view synthesis \cite{watson2022novel, gu2023nerfdiff, tseng2023consistent} and 3D domains \cite{Zhou_2023_CVPR, anciukevivcius2023renderdiffusion}.

\subsection{Neural  3D Representations}

Traditionally, the representation of 3D models and shapes has predominantly relied on explicit methods such as voxels, meshes, and point clouds. While these explicit representations don't necessitate intricate rendering steps, they come with their inherent limitations, including issues such as memory complexity due to the 3-dimensional data increasing in an order of 3. In response to these challenges, an alternative approach involves representing 3D shapes implicitly using neural networks, which have memory complexities correlated with the parameter number of the neural networks. Two notable techniques in this domain are DeepSDF and DeepLS, wherein models are stored as learned signed distance functions; signed distance functions where the function returns the distance of each point to the inside of the object\cite{chabra2020deep, park2019deepsdf}. This approach has been adopted in various other methods, including MonoSDF, InstantNGP, and VolSDF, \cite{yu2022monosdf, mueller2022instant, NEURIPS2021_25e2a30f}.

Similarly, neural radiance fields (NeRF) have gained serious attention to model real-life scenes. NeRFs are shallow multilayer perceptrons that learn to output the color and density value of a pixel corresponding to the input 5D camera positions\cite{mildenhall2020nerf}. A more recent addition to the repertoire of 3D scene representations is Gaussian splatting, where the 3D model is represented as a collection of Gaussian distributions with learnable mean and standard deviations \cite{kerbl20233d}. Relative to the other methods, Gaussian splats are easier to render as these models are explicit, therefore they do not require any postprocessing such as marching cubes algorithm to convert from implicit representations.

\section{Material and Method}
This section will first present material about the International Union for Conservation of Nature (IUCN) and the Red List of Threatened Species. Next, introduce the overall working method of DreamGaussian. Firstly, 3D Gaussian Splatting will be presented. Then, the DreamGaussian method along with the adaptation of Gaussian Splatting in generative settings will be explained.
\subsection{Material}
The Red List of Threatened Species, created and maintained by the IUCN, is a globally recognized and widely used resource for assessing the conservation status of various species. The primary goal of the IUCN Red List is to provide an up-to-date and comprehensive assessment of the global extinction risk faced by animal, plant, and fungal species. It serves as a tool for identifying species that are at risk of extinction, helping prioritize conservation efforts and resources. The Red List uses a set of standardized criteria to evaluate the conservation status of species. These criteria include factors such as population size, rate of decline, distribution, and the degree of threat from various pressures like habitat loss, exploitation, disease, and climate change. Species assessed by the Red List are categorized into specific threat categories. These categories include: extinct, extinct in the wild, critically endangered, endangered, vulnerable, near threatened, least concern, data deficient and not evaluated. The IUCN Red List stands as a critical resource in the ongoing effort to monitor and conserve global biodiversity, providing valuable insights into the status of species and informing conservation strategies on a global scale.
\subsection{Method}
\subsubsection{3D Gaussian Splatting} \label{3dgs}
3D Gaussian Splatting \cite{KKLD23} is basically a rasterization technique that reconstructs 3D scenes from a few images that are taken from different views of an object or a scene. It represents the scenes using Gaussians. For the representation, it uses 4 parameters; position (x,y,z), covariance, RGB, and alpha ($\alpha$) values. There are numerous reasons why 3D Gaussians are better than traditional mesh-based methods. Generally, the most prevalent way of representing 3D shapes is explicit representation such as point clouds, meshes, or voxels. Because it is easy to retrieve features from these data structures. Moreover, they allow fast rendering and training. Another way is implicit representations\cite{mildenhall2020nerf} which includes stochastic sampling and it has the advantage of better optimization because of its continuity. However, it is computationally expensive for rendering. 3D Gaussian splatting brings the best of both worlds. As we mentioned above the representation of Gaussians, their optimization is easy because they are differentiable (advantage of implicit representation) and are sufficient for rendering since they are like meshes (advantage of explicit representation). Therefore, 3D Gaussians accomplish fast rendering and continuity while being differentiable.

The next step is to determine the four parameters described in the first paragraph. First of all, the covariance matrix regulates the spread of the distribution (scaling and stretching) and must be symmetric and positive definite. Secondly, color is represented by spherical harmonic (SH) coefficients which are a set of functions that encode data. For 2D, we can define functions dependent on vertex coordinates, and changing the values represents any color. This way, 2D images can be represented by equations instead of storing the RGB values. So, with the help of SH, this can be applied to 3D. By defining SH harmonics, colors can be stored at specific points and calculated by spherical harmonics functions. Basically, SH is a collection of functions that takes inputs; polar angle ($\theta$), distance from the center (\textit{r}), and azimuth angle ($\phi$). The resultant is an output value at a point on a sphere's surface. Consequently, spherical harmonics functions give the output at specific points on the surface of a sphere.
\begin{figure*}[t]
	\centering
	\includegraphics[width=1.0\textwidth]{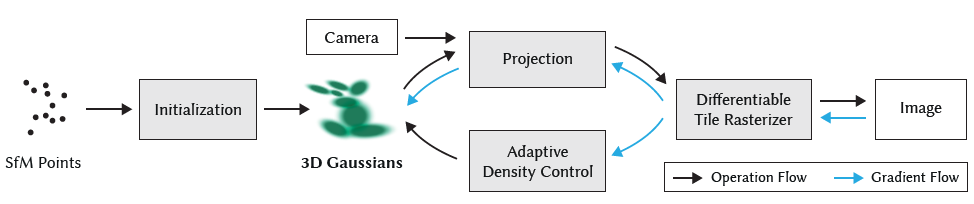}
	\caption{Overall pipeline of 3D Gaussian Splatting \cite{KKLD23}}
	\label{fig:pipeline}
\end{figure*}
Now, the step before optimization is to initialize the parameters. Firstly, a pipeline called COLMAP \cite{schoenberger2016sfm} utilizes the Structure from Motion (SfM) algorithm to generate a point cloud. It generates sparse and/or dense point clouds using multi-view images of the same object. In the Gaussian Splatting, these points are used as the means of the 3D Gaussians (for each point). The covariance matrix ($\Sigma$) consists of a rotation matrix (R) and a scale matrix (S).
\begin{equation}
    \Sigma = RSS^TR^T
\end{equation}
Initial values of the scale matrix (3x3) are obtained from the log transformation of the square root of the distances to each point in the point cloud. The rotation matrix is represented as quaternions (4x1) and initialized with the first value as 1 and the rest are 0. Then, it is converted to a rotation matrix (3x3). Color values are initialized from the values belonging to the points obtained by SfM. Transparency ($\alpha$) is also set with a predefined value by the authors.

\begin{figure*}[h]
	\centering
	\includegraphics[width=0.7\textwidth]{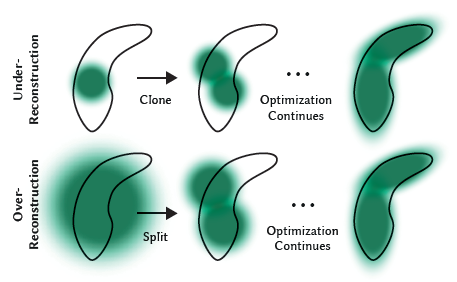}
	\caption{Adaptive Gaussian Densification \cite{KKLD23}}
	\label{fig:densification}
\end{figure*}

The training process utilizes Stochastic Gradient Descent techniques for optimization. The simplest explanation of the training procedure starts with rasterizing the Gaussian to an image with the help of differentiable Gaussian rasterization. This function takes the camera pose, width, and height of the image along with the initialized parameters mentioned above. It includes projecting the Gaussians into 2D, sorting them by view space depth, and finally iterating each Gaussian for each pixel and then blending them. Finally, returns an image which is then fed into the loss function with the ground truth image. The loss function is:
\begin{equation}
    \mathcal{L} = {(1-\lambda)}\mathcal{L}_1 + \lambda\mathcal{L}_{D-SSIM}
\end{equation}
There is also another step in the optimization part which is called Adaptive Control of Gaussians. This part densifies or prunes the Gaussians. Firstly, if alpha is lower than a specific threshold (0.005) it prunes the corresponding Gaussians. The densification part is for over or under-reconstruction.
Under-reconstruction occurs when there aren't adequate Gaussians for the space. The solution to fit the space is to clone the Gaussians along the gradient direction. On the other hand, over-reconstruction is the opposite; when a Gaussian is larger than the desired space. This approach splits the single Gaussian into smaller ones and places them according to the probability density values of the initial one. The visual examples of both processes are shown in Figure \ref{fig:densification}

In summary, the method starts by retrieving the point clouds from the SfM algorithm. The next step is to initialize 3D Gaussian values with the help of the point clouds. Then, these are projected into 2D Gaussians and passed through rasterization to obtain image predictions. And, these images go into loss function with ground truths and backpropagated for the value update. Finally, during this process, adaptive density control is applied systematically.

\subsubsection{DreamGaussian}
DreamGaussian \cite{tang2023dreamgaussian} is a novel method for image-to-3D and text-to-3D generation. The key idea in DreamGaussian is using 3D Gaussian Splatting and combining it with sophisticated mesh extraction and texture refinement algorithms. It has two stages; the first one is mesh extraction and the last one is mesh and texture refinement. It starts with 3D Gaussian splitting parameter initialization as described in Section \ref{3dgs}. It is worth noting that this method utilizes 3D Gaussian Splatting in generative settings. However, in DreamGaussian, these are initialized randomly inside a sphere. Densification is also performed periodically. Authors used Score Distillation Sampling (SDS) \cite{poole2022dreamfusion} for the optimization of 3D Gaussians. Each step includes the following processes; random camera pose sampling, rendering the RGB image, and transparency of the corresponding view. Random noise is determined by decreasing the timestep linearly and adding to the rendered image. And finally, the SDS denoising step is guided by 2D diffusion priors which are eventually back-propagated to the Gaussians.

\begin{figure*}[h]
	\centering
	\includegraphics[width=0.6\textwidth]{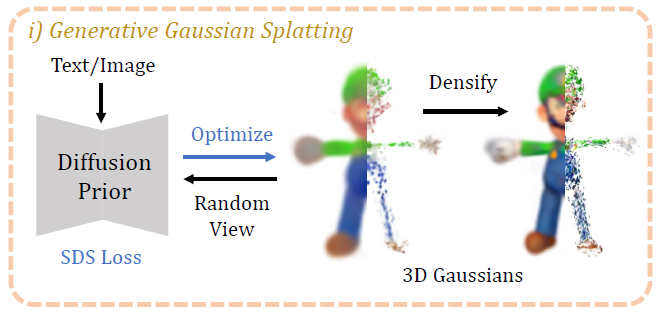}
	\caption{Generative Gaussian Splatting Stage \cite{tang2023dreamgaussian}}
	\label{fig:gengs}
\end{figure*}

For the image-to-3D, the input is an image and a foreground mask. Mesh extraction is done from 3D Gaussians with an algorithm called local density query and back-projected color. In this part, the authors take advantage of pruning and/or splitting properties of 3D Gaussians to accomplish block-wise density queries. The first step is to divide 3D space into $16^3$ blocks. The next step involves pruning the Gaussians with the centers outside each corresponding block. Then, another dense query into $8^3$ for each local block is performed. The result can be obtained for each grid position;
\begin{equation}\label{densityeq}
    d(x) = \sum\limits_{i=1}{a_i}{exp(-1/2(x-x_i)^T\Sigma_{i}^{-1}(x-x_i))}
\end{equation}
In the Equation \ref{densityeq}, $\Sigma_{i}$ is the covariance matrix as it is described in \ref{3dgs}. Then a threshold is set which was found empirically for extracting the mesh surface using Marching Cubes\cite{Lorensen1987MarchingCA}. 

\begin{figure*}[h]
	\centering
	\includegraphics[width=0.4\textwidth]{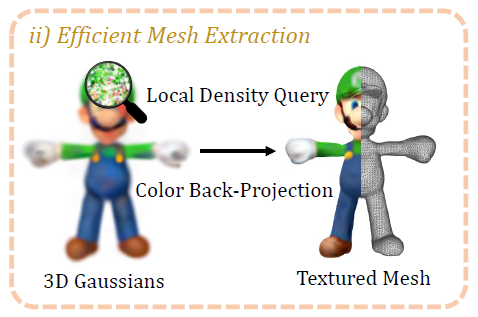}
	\caption{Mesh Extraction \cite{tang2023dreamgaussian}}
	\label{fig:meshextr}
\end{figure*}

\begin{figure*}[h]
	\centering
	\includegraphics[width=1\textwidth]{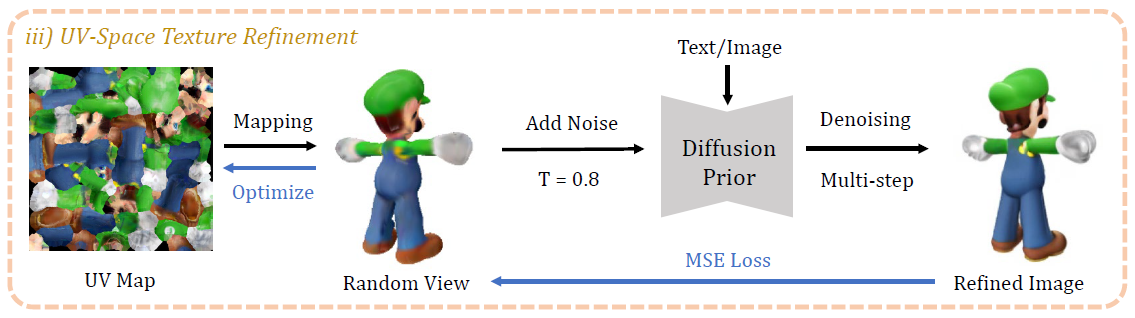}
	\caption{UV-Texture Refinement \cite{tang2023dreamgaussian}}
	\label{fig:meshrefine}
\end{figure*}

After obtaining the mesh, the next step is to map the image (rendered) to the mesh surface as texture. It is achieved by selecting 8 azimuth and 3 elevation angles uniformly along with the top and bottom views to render the image. Based on the UV coordinate, each pixel from the images is back-projected and the final result is used for the next fine-tuning mesh texture process. The necessity of the second stage comes from the vagueness of the SDS optimization. This ambiguity results in a blurry texture and this stage enhances the texture image. Authors are inspired by SDEdit\cite{meng2022sdedit} to overcome this problem. Process for this stage is shown in Figure \ref{fig:meshrefine}
The rendering process of a blurry image from an arbitrary camera pose can be accomplished by having an initialization texture. The refined image can be obtained by perturbing the image with random noise and applying a multi-step denoising process. Then, the refined image is used for optimization of the texture through pixel-wise MSE loss

\subsubsection{Experimental Setup}
This section begins with experimental setup details that explain inputs and parameters. The visual results of generated 3D models will be presented along with the effects of the given input parameter in \ref{sec:res} Results and Discussion Section.

\begin{figure*}[h]
	\centering
	\includegraphics[width=0.4\textwidth]{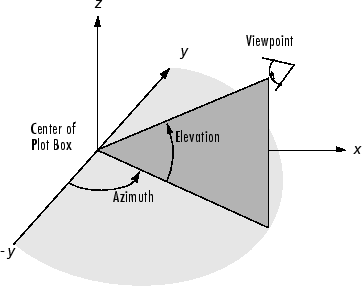}
	\caption{Elevation Angle Definition}
	\label{fig:elevation}
\end{figure*}

We conduct comprehensive experiments using various elevation angles as input to the DreamGaussian method. Elevation angle ($\varepsilon$) is the angle between the object and the observer's horizontal plane as seen in Figure \ref{fig:elevation}
We've chosen 3 species as input images with 7 elevation angles; -30, -20, -10, 0, 10, 20, and 30. The results will include stage 2 of DreamGaussian too. All of the parameters are set to default except the elevation angle.

\section{Results and Discussion}
\label{sec:res}
We conduct experiments on 3 different species which have different physical attributes. Since DreamGaussian method requires only a single image for 3D generation, we've chosen only one input image for each animal. The input images are shown in below.

\begin{figure}[h]
     \centering
     \begin{subfigure}[h]{0.33\textwidth}
         \centering
         \includegraphics[width=5cm, height=5cm]{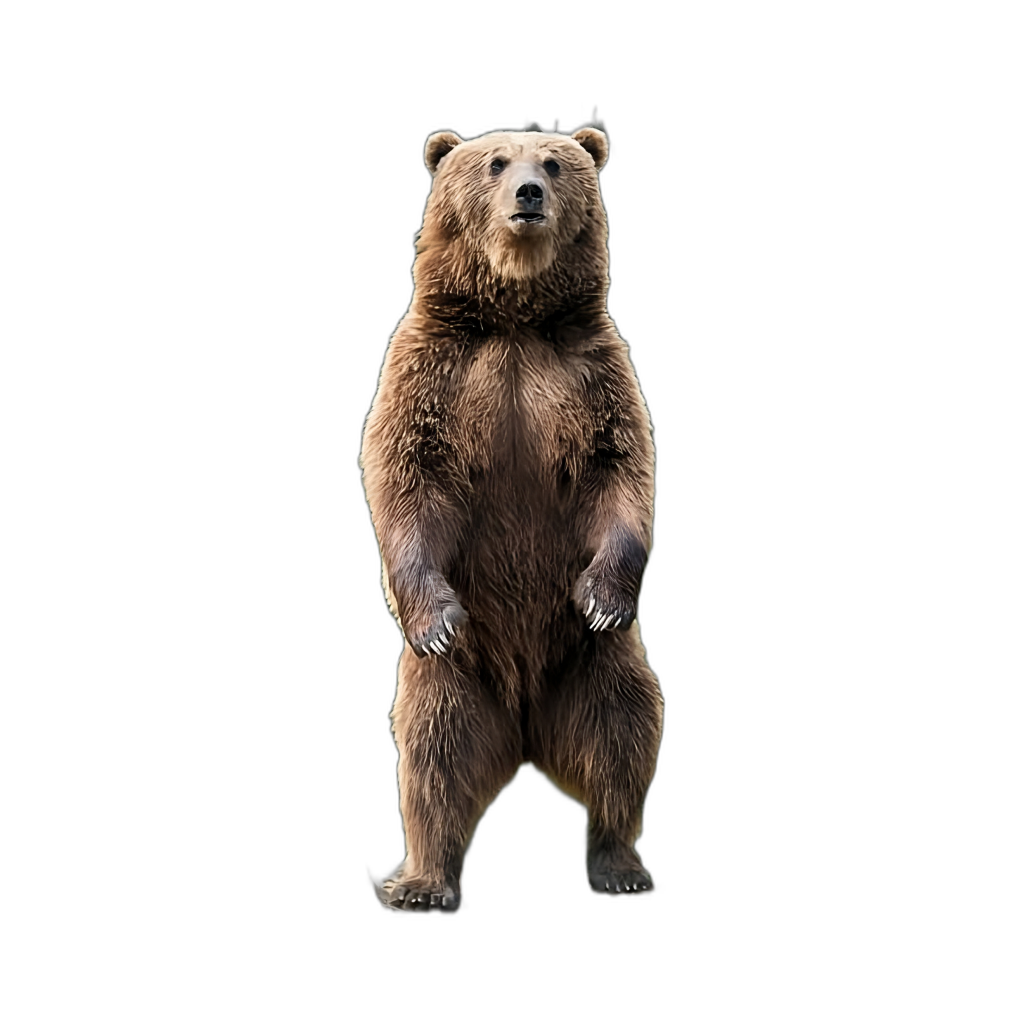}
         \label{fig:bear}
     \end{subfigure}
     \begin{subfigure}[h]{0.33\textwidth}
         \centering
         \includegraphics[width=5cm, height=5cm]{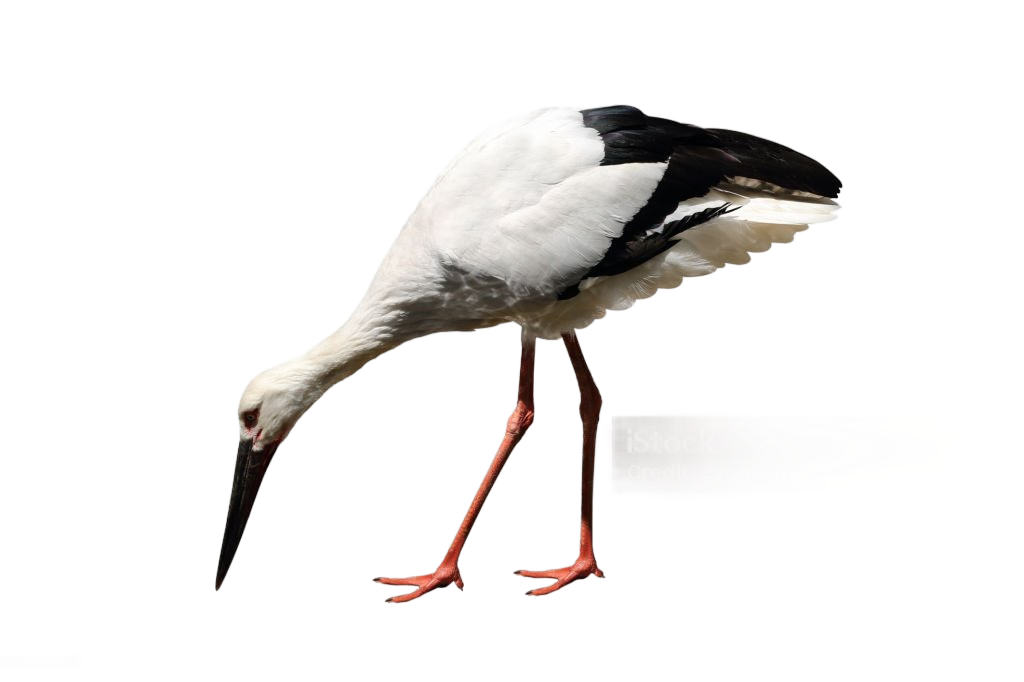}
         \label{fig:stork}
     \end{subfigure}
     \begin{subfigure}[h]{0.33\textwidth}
         \centering
         \includegraphics[width=5cm, height=5cm]{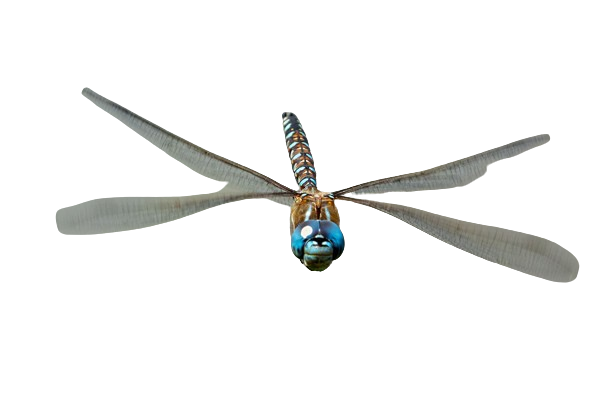}
         \label{fig:dragonfly}
     \end{subfigure}
        \caption{Input images (a) Grizzly Bear, (b) Oriental White Stork and (c) Dragonfly}
        \label{fig:inputs}
\end{figure}
Model visualizations are performed in MeshLab, an open source software and designed for the purpose of processing and modifying 3D triangular meshes. For each input image, we used 7 elevation angles and we captured the images of the 3D model from 3 different views. Generally, if the object in the image appears in a top-down view, we need to increase the elevation angle input (positive). Otherwise, the input should be negative. Commonly, the input angle ranges from [-30, 30].
\begin{figure}
    \centering
    \begin{tabular}{cccccccc}
        &$\varepsilon$=-30  &$\varepsilon$=-20  &$\varepsilon$=-10  &$\varepsilon$=0    &$\varepsilon$=10   &$\varepsilon$=20   &$\varepsilon$=30\\ 
        & \includegraphics[width=1.8cm, height=3.2cm]{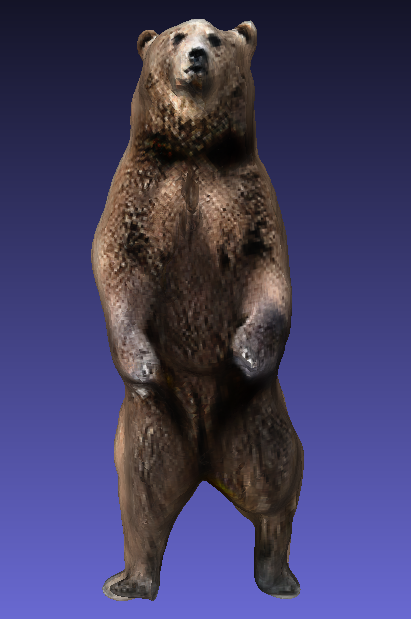} 
        & \includegraphics[width=1.8cm, height=3.2cm]{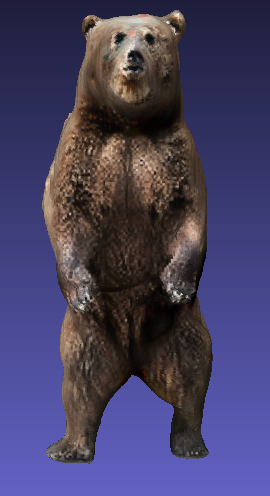} 
        & \includegraphics[width=1.8cm, height=3.2cm]{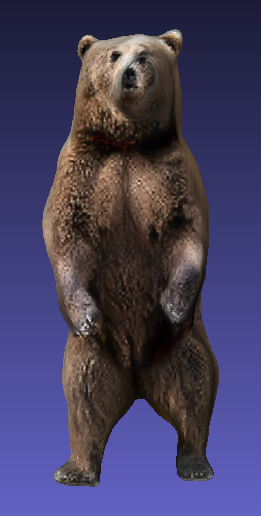} 
        & \includegraphics[width=1.8cm, height=3.2cm]{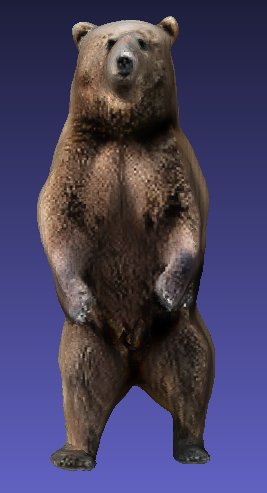} 
        & \includegraphics[width=1.8cm, height=3.2cm]{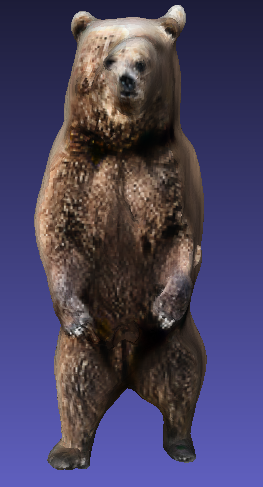} 
        & \includegraphics[width=1.8cm, height=3.2cm]{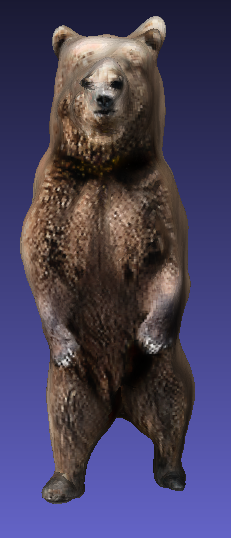} 
        & \includegraphics[width=1.8cm, height=3.2cm]{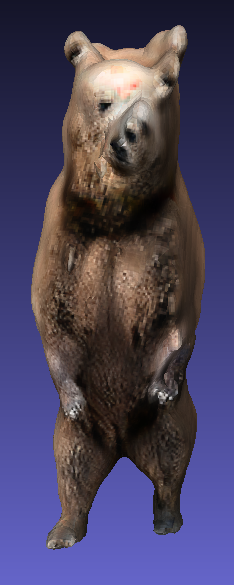}\\
    
        & \includegraphics[width=1.8cm, height=3.2cm]{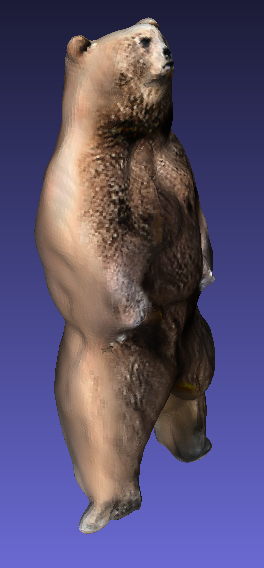} 
        & \includegraphics[width=1.8cm, height=3.2cm]{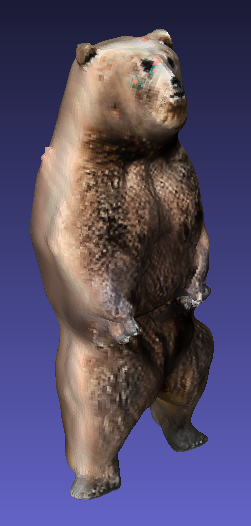} 
        & \includegraphics[width=1.8cm, height=3.2cm]{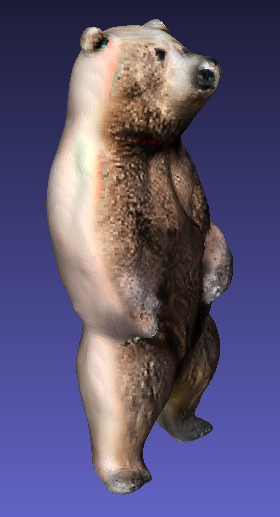} 
        & \includegraphics[width=1.8cm, height=3.2cm]{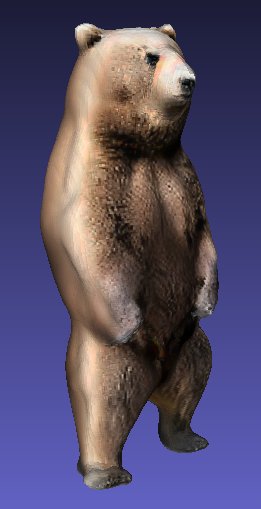} 
        & \includegraphics[width=1.8cm, height=3.2cm]{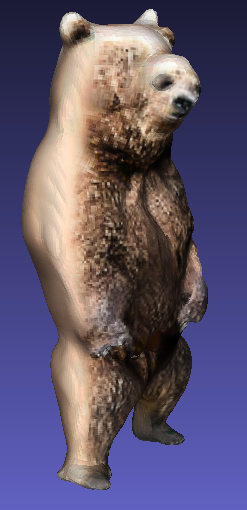} 
        & \includegraphics[width=1.8cm, height=3.2cm]{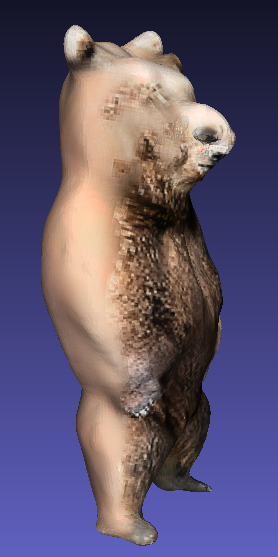} 
        & \includegraphics[width=1.8cm, height=3.2cm]{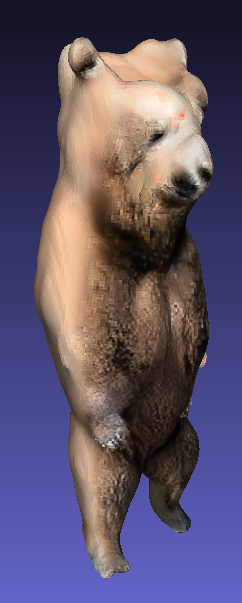}\\
        
        & \includegraphics[width=1.8cm, height=3.2cm]{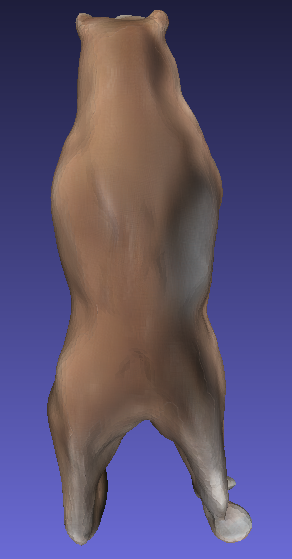} 
        & \includegraphics[width=1.8cm, height=3.2cm]{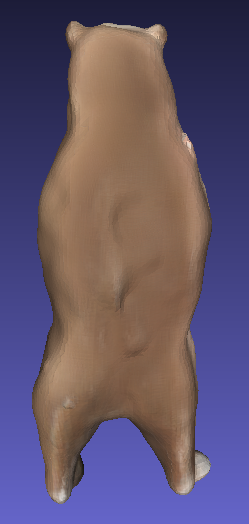} 
        & \includegraphics[width=1.8cm, height=3.2cm]{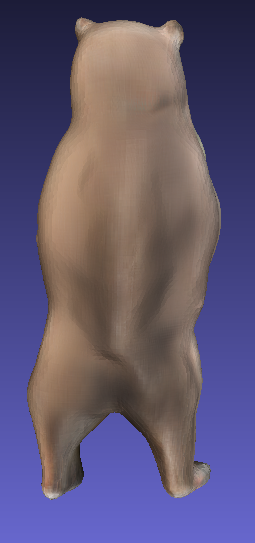} 
        & \includegraphics[width=1.8cm, height=3.2cm]{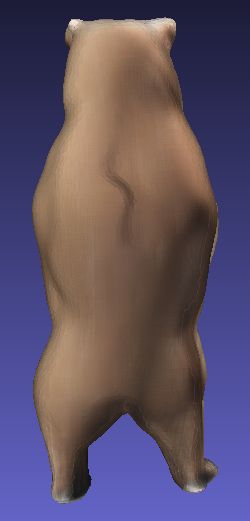} 
        & \includegraphics[width=1.8cm, height=3.2cm]{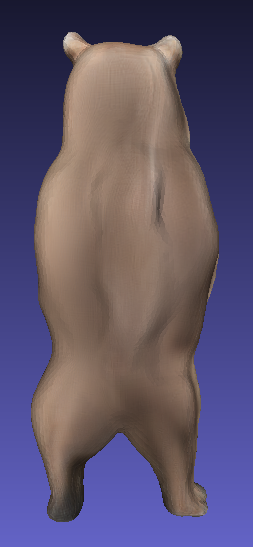} 
        & \includegraphics[width=1.8cm, height=3.2cm]{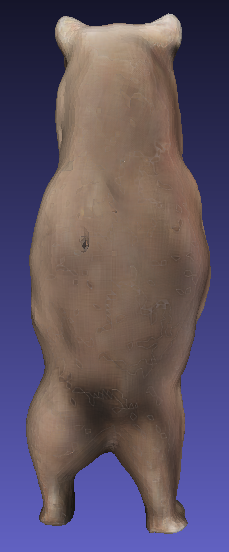} 
        & \includegraphics[width=1.8cm, height=3.2cm]{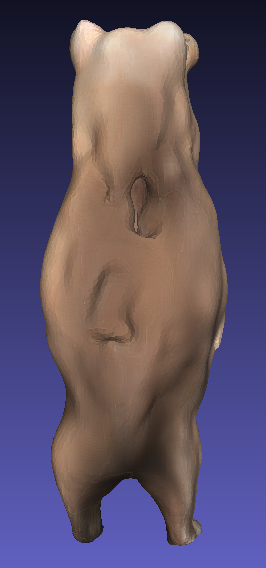}
        
    \end{tabular}
    \caption{3D Outputs of Grizzly Bear with specified elevation angles (Each column belongs to the same output but different views)}
    \label{bear_res}
\end{figure}
As it can be seen in Figure \ref{bear_res}, the 3D results of the bear become worse for all the elevation angles except 0. This is because the bear in the input image appears to be looking directly towards the camera. Therefore, the elevation angle is approximately around 0. Wrong elevation input causes poor mesh generation. Since the mesh is not aligned with the original physical structure of the bear, texture mapping is also not successful.
\begin{figure}[h]
    \centering
    \begin{tabular}{cccccccc}
        &$\varepsilon$=-30&$\varepsilon$=-20&$\varepsilon$=-10&$\varepsilon$=0&$\varepsilon$=10&$\varepsilon$=20&$\varepsilon$=30\\
        & \includegraphics[width=1.9cm, height=2.0cm]{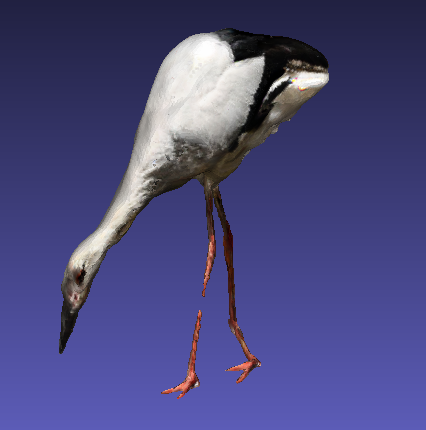}
        & \includegraphics[width=1.9cm, height=2.0cm]{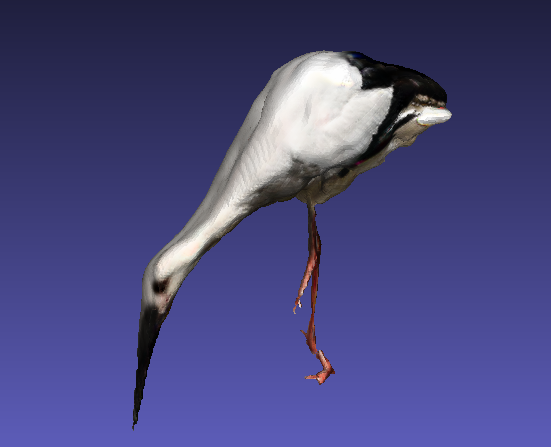} 
        & \includegraphics[width=1.9cm, height=2.0cm]{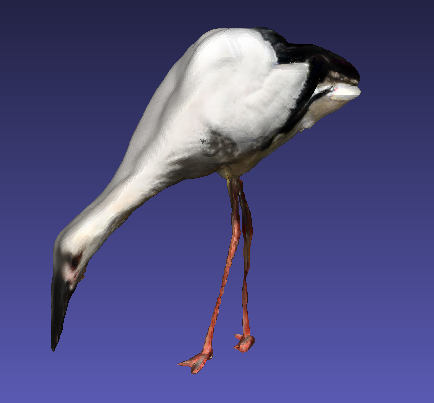} 
        & \includegraphics[width=1.9cm, height=2.0cm]{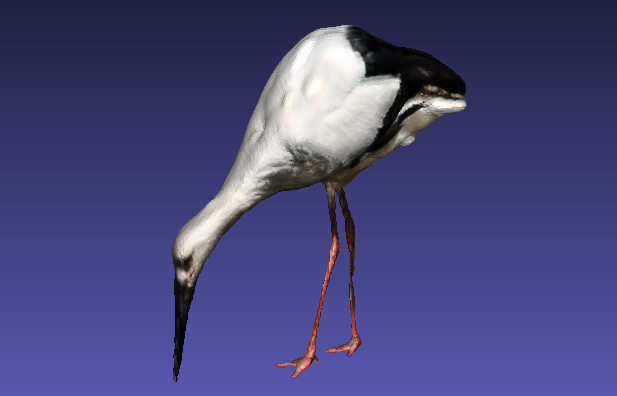} 
        & \includegraphics[width=1.9cm, height=2.0cm]{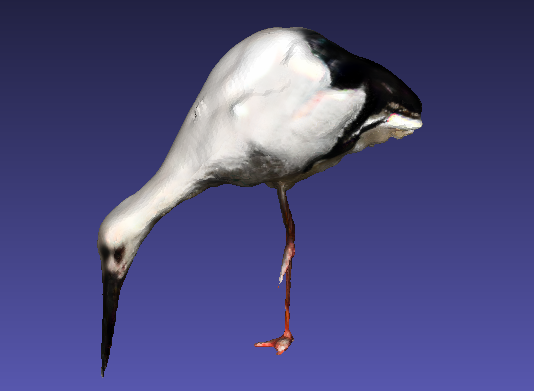} 
        & \includegraphics[width=1.9cm, height=2.0cm]{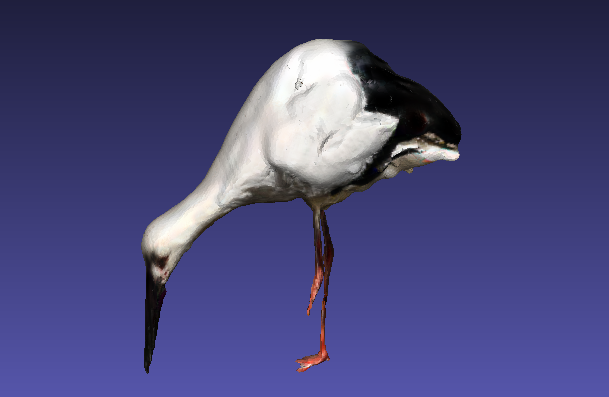} 
        & \includegraphics[width=1.9cm, height=2.0cm]{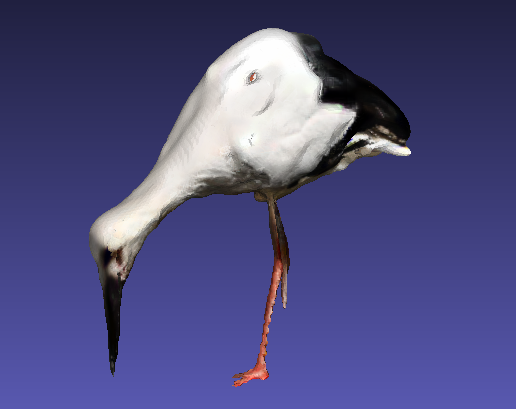}\\
    
        & \includegraphics[width=1.9cm, height=2.0cm]{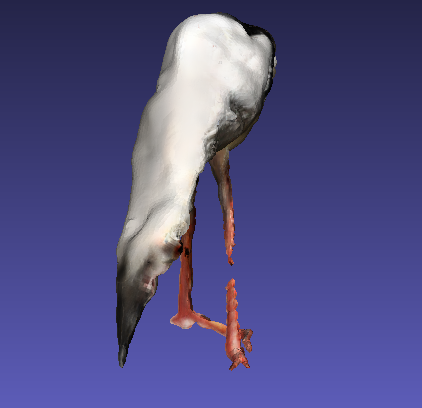} 
        & \includegraphics[width=1.9cm, height=2.0cm]{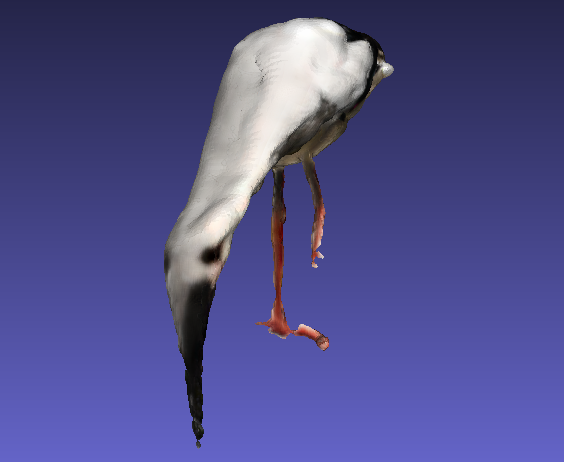} 
        & \includegraphics[width=1.9cm, height=2.0cm]{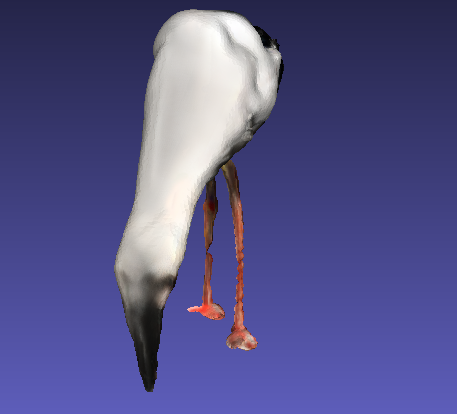} 
        & \includegraphics[width=1.9cm, height=2.0cm]{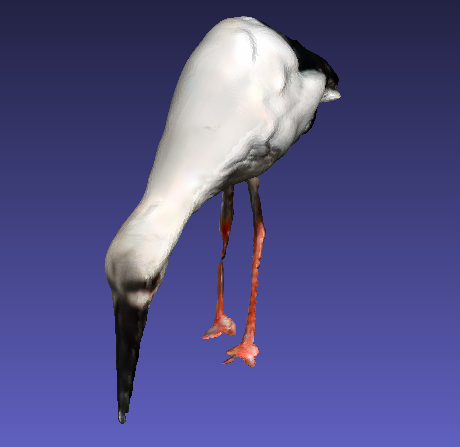} 
        & \includegraphics[width=1.9cm, height=2.0cm]{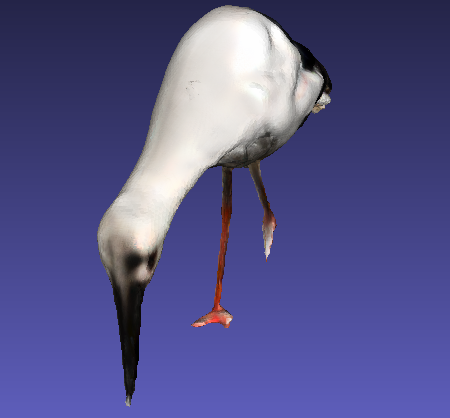} 
        & \includegraphics[width=1.9cm, height=2.0cm]{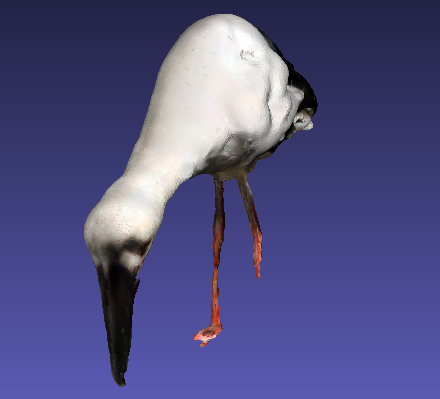} 
        & \includegraphics[width=1.9cm, height=2.0cm]{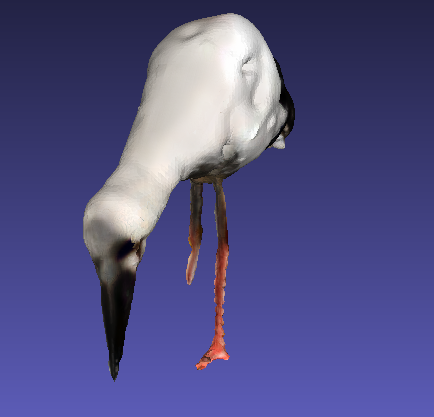}\\
        
        & \includegraphics[width=1.9cm, height=2.0cm]{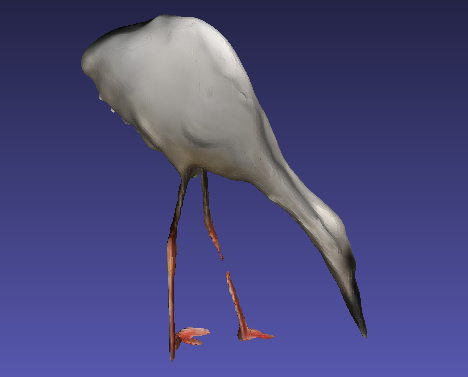} 
        & \includegraphics[width=1.9cm, height=2.0cm]{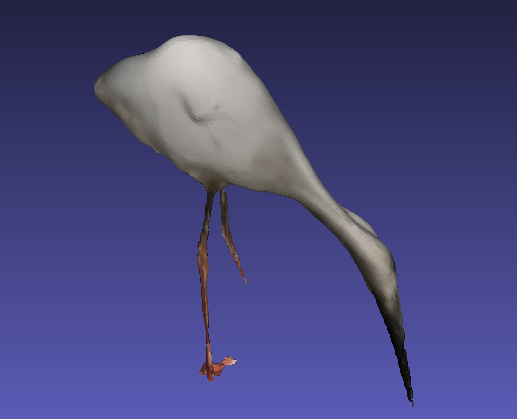} 
        & \includegraphics[width=1.9cm, height=2.0cm]{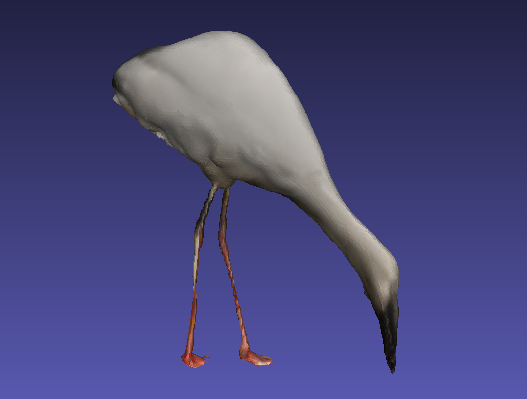} 
        & \includegraphics[width=1.9cm, height=2.0cm]{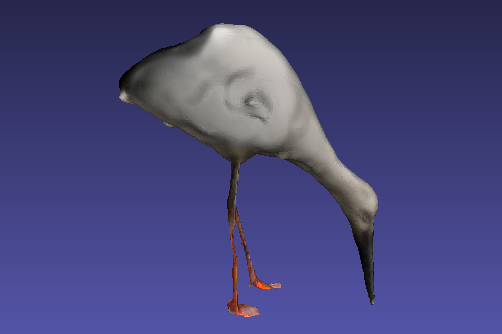} 
        & \includegraphics[width=1.9cm, height=2.0cm]{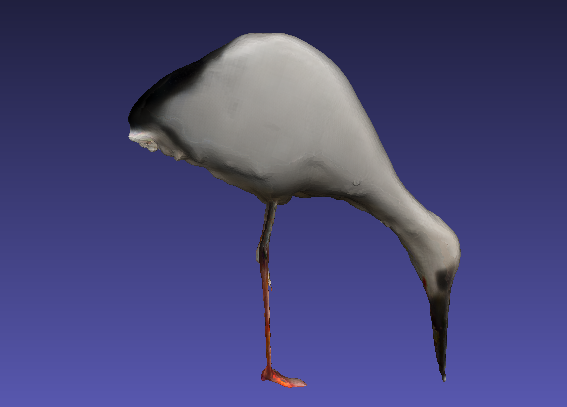} 
        & \includegraphics[width=1.9cm, height=2.0cm]{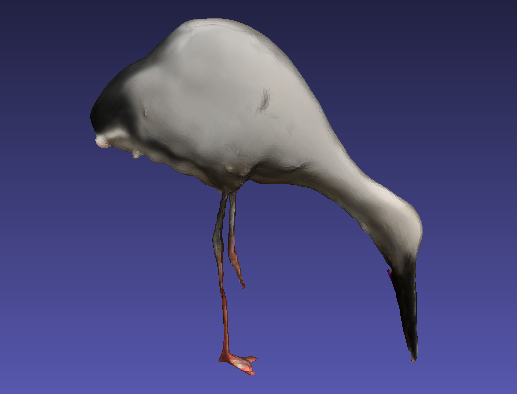} 
        & \includegraphics[width=1.9cm, height=2.0cm]{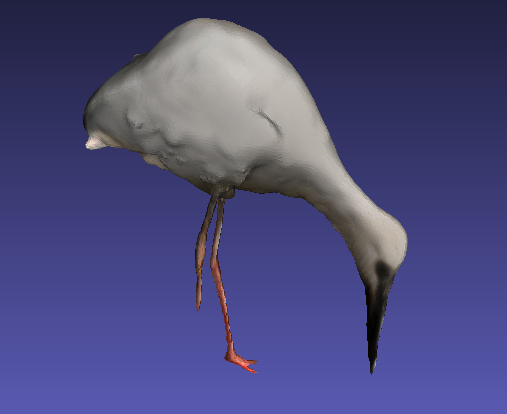}
        
    \end{tabular}
    \caption{3D Outputs of Oriental White Stork with specified elevation angles}
    \label{fig:stork_res}
\end{figure}
 The result of the stork also shows the same behavior in Figure \ref{fig:stork_res}. For the edge values of the input angle, the result is not acceptable. For example, for the values $\varepsilon$=[0,-10], the 3D model seems fine. Especially for the angle 0, the best output is produced since -10 has some unusual body shape around the head. However, the other inputs produced discontinuities in the mesh around the leg parts.
 \begin{figure}[h]
    \centering
    \begin{tabular}{cccccccc}
        &$\varepsilon$=-30&$\varepsilon$=-20&$\varepsilon$=-10&$\varepsilon$=0&$\varepsilon$=10&$\varepsilon$=20&$\varepsilon$=30\\
        & \includegraphics[width=1.9cm, height=1.6cm]{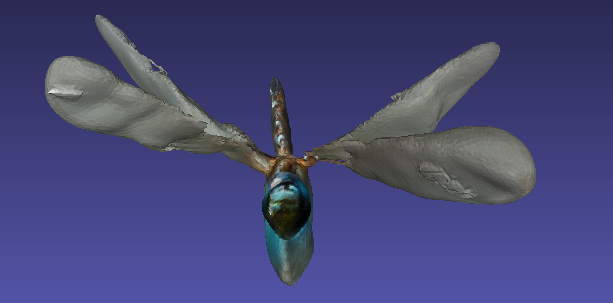}
        & \includegraphics[width=1.9cm, height=1.6cm]{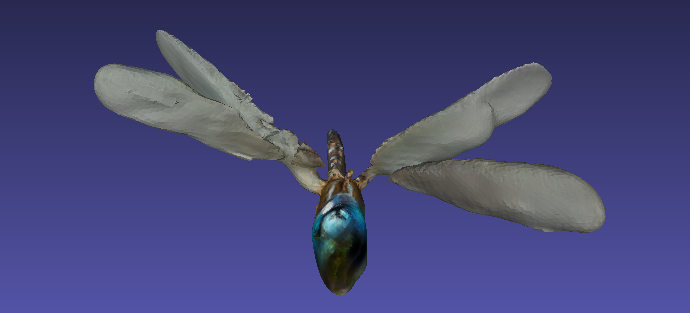} 
        & \includegraphics[width=1.9cm, height=1.6cm]{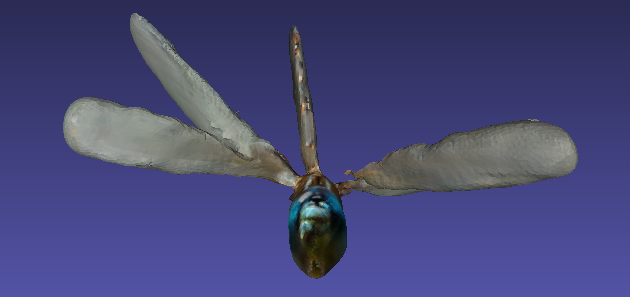} 
        & \includegraphics[width=1.9cm, height=1.6cm]{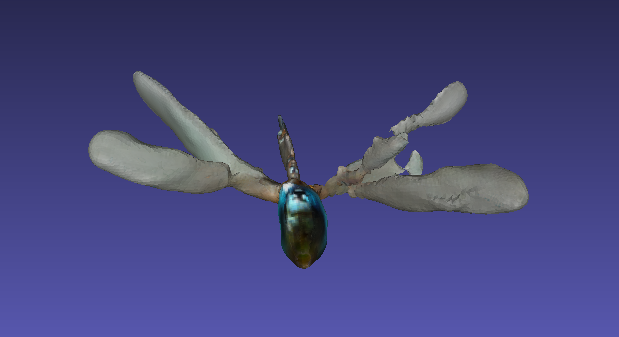} 
        & \includegraphics[width=1.9cm, height=1.6cm]{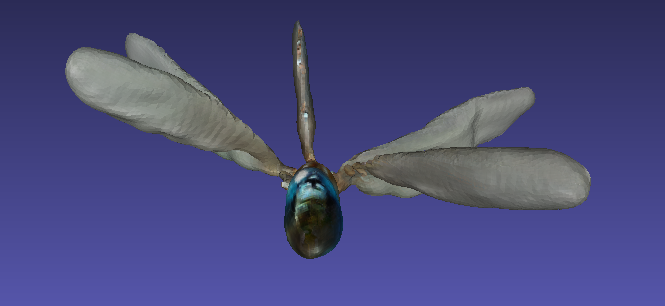} 
        & \includegraphics[width=1.9cm, height=1.6cm]{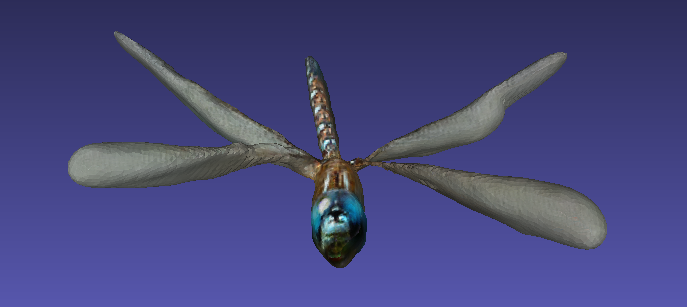} 
        & \includegraphics[width=1.9cm, height=1.6cm]{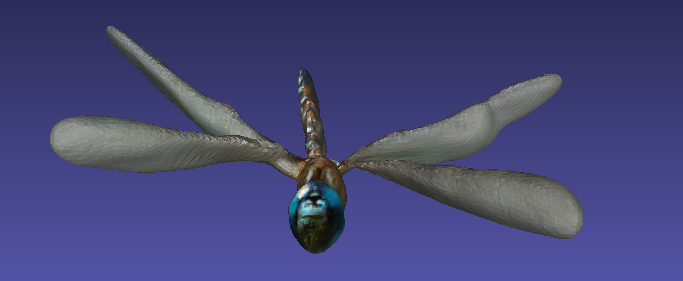}\\
    
        & \includegraphics[width=1.9cm, height=1.6cm]{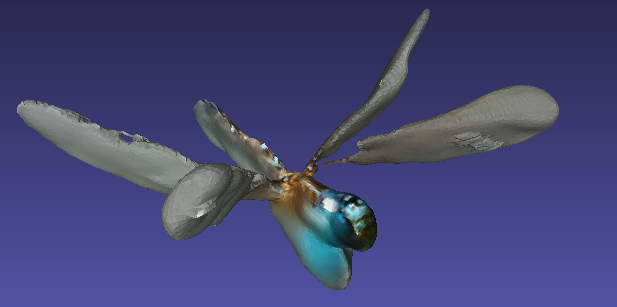} 
        & \includegraphics[width=1.9cm, height=1.6cm]{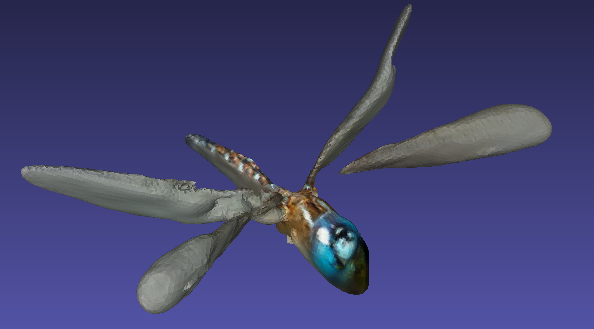} 
        & \includegraphics[width=1.9cm, height=1.6cm]{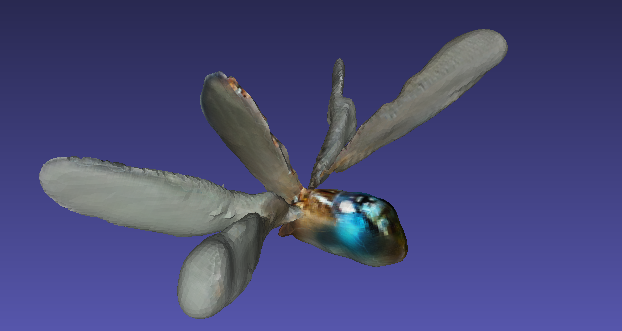} 
        & \includegraphics[width=1.9cm, height=1.6cm]{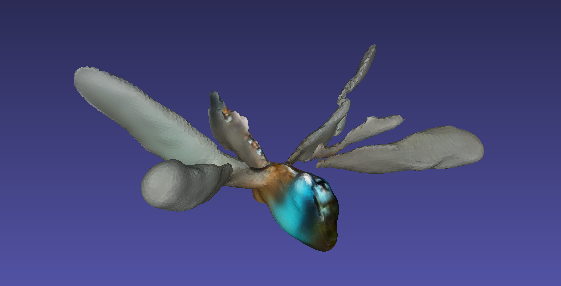} 
        & \includegraphics[width=1.9cm, height=1.6cm]{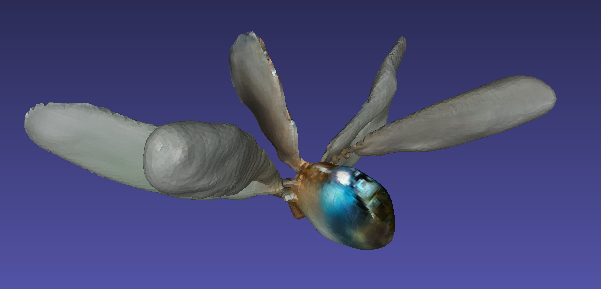} 
        & \includegraphics[width=1.9cm, height=1.6cm]{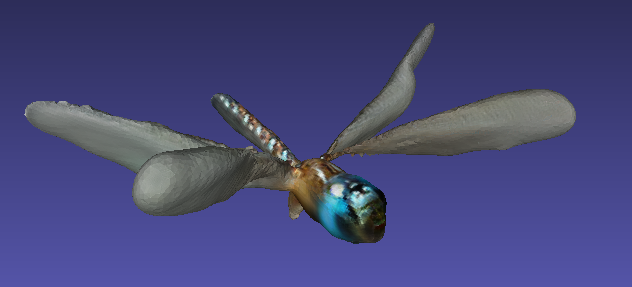} 
        & \includegraphics[width=1.9cm, height=1.6cm]{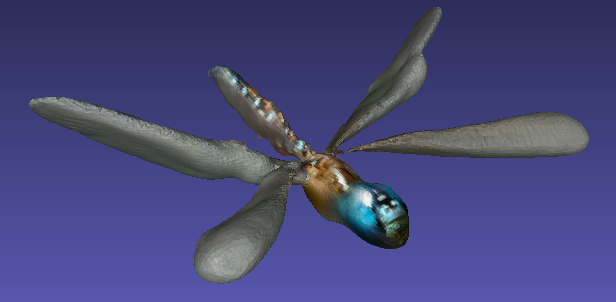}\\
        
        & \includegraphics[width=1.9cm, height=1.6cm]{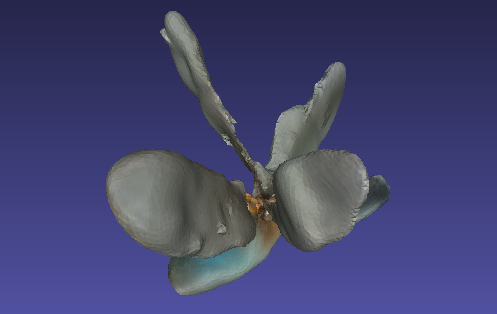} 
        & \includegraphics[width=1.9cm, height=1.6cm]{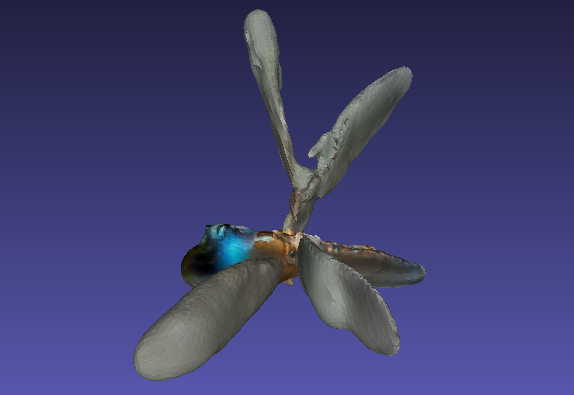} 
        & \includegraphics[width=1.9cm, height=1.6cm]{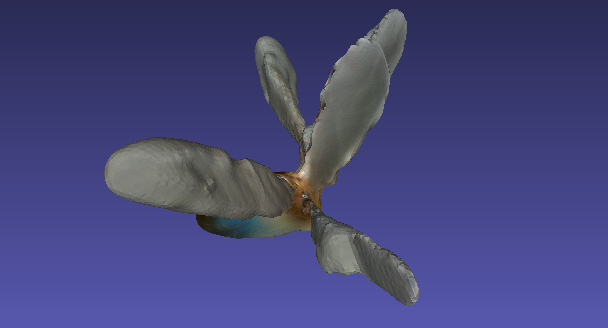} 
        & \includegraphics[width=1.9cm, height=1.6cm]{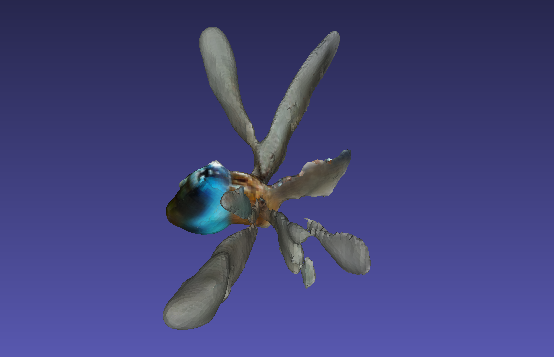} 
        & \includegraphics[width=1.9cm, height=1.6cm]{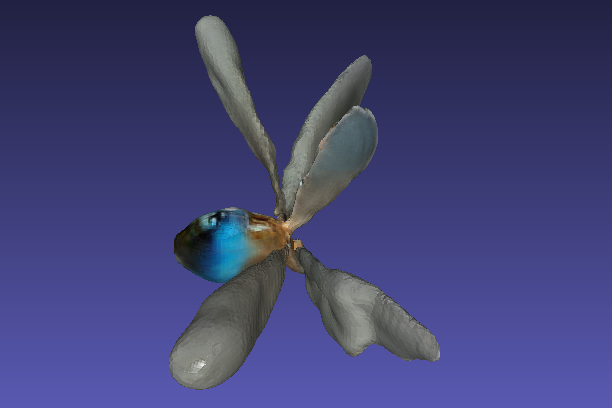} 
        & \includegraphics[width=1.9cm, height=1.6cm]{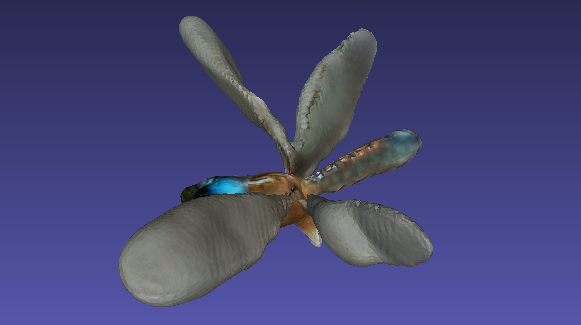} 
        & \includegraphics[width=1.9cm, height=1.6cm]{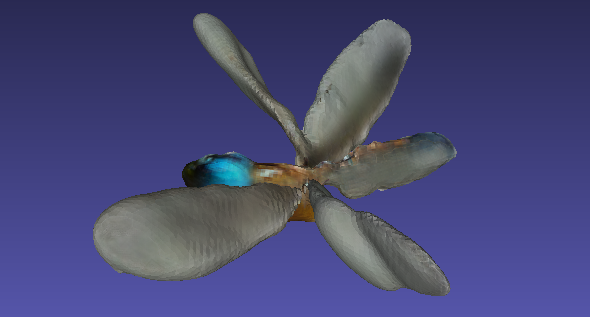}
        
    \end{tabular}
    \caption{3D Outputs of Dragonfly with specified elevation angles}
    \label{fig:dragonfly_res}
\end{figure}
The results for the Dragonfly are different from the other two inputs. As can be observed in Figure \ref{fig:dragonfly_res}, the best result is when the elevation angle is 30. This validates our findings because if we look at the input image, the camera looks from the top-down perspective to the Dragonfly. That is why when the input is around positive 30, the output is better.

\section{Conclusion}
In conclusion, the threat of extinction faced by many species highlights the importance of preserving biodiversity. However, due to the rarity of endangered animals, there is limited data available, making it challenging to apply data requiring generative AI methods to this domain. This study aimed to investigate the feasibility of generating consistent and real-like 3D models of endangered animals using limited data by utilizing zero-shot stable diffusion models. The focus of this study was on DreamGaussian, a novel framework that utilizes Generative Gaussian Splatting along with novel mesh extraction and refinement algorithms.

Through a comprehensive analysis, we observed that giving a correct elevation angle with the input image significantly affects the result of the generated 3D model. Overall, DreamGaussian is a fairly good method for generating 3D models from single images. In DreamGaussian, the elevation angle is set to 0 as default. This leads to unexpected degradation in results if the object in the given image is from a very different angle. It distorts both the texture and structure of the generated mesh. However, if we provide a reliable elevation angle with the input image, it's observed that the outcome will be better. 

On the other hand, apart from the elevation angle, the 3D model also has some issues with the texture of unseen parts. There is still room for improvement. 

We hope that this study will be influential in the usability of AI to preserve endangered animals and will be a step forward in our goal of obtaining a model that can output biologically consistent 3D models via small samples.

\bibliographystyle{unsrt}
\bibliography{references}

\end{document}